\documentclass{article}

\usepackage{microtype}
\usepackage{graphicx}
\usepackage{subcaption}
\usepackage{booktabs} 
\usepackage{xspace}
\usepackage{makecell}
\usepackage{amsfonts}
\usepackage[T1]{fontenc}

\usepackage{xcolor}

\usepackage{hyperref}

\usepackage[accepted]{icml2018}

\icmltitlerunning{The Single Headed Attention RNN}

\newcommand{\enwik}{\texttt{enwik8}\xspace}

\begin{document}

\twocolumn[
\icmltitle{Single Headed Attention RNN: Stop Thinking With Your Head}



\icmlsetsymbol{equal}{*}

\begin{icmlauthorlist}
\icmlauthor{Stephen Merity}{to}
\end{icmlauthorlist}

\icmlaffiliation{to}{$\frac{d}{dx}$ Times Labs}
\icmlcorrespondingauthor{Stephen Merity}{smerity@smerity.com}
\icmlkeywords{language modeling, LSTM, QRNN, RNN}

\vskip 0.3in
]



\printAffiliationsAndNotice{}  

\begin{abstract}
The leading approaches in language modeling are all obsessed with TV shows of my youth - namely Transformers and Sesame Street.
Transformers this, Transformers that, and over here a bonfire worth of GPU-TPU-neuromorphic wafer scale silicon.
We opt for the lazy path of old and proven techniques with a fancy crypto\footnote{Crypto here expands to cryptography though I'd sure be happy to receive whatever fancy AI singularity Universal Basic Compute Income coin you're willing to give obviously.} inspired acronym: the Single Headed Attention RNN (SHA-RNN).
The author's lone goal is to show that the entire field might have evolved a different direction if we had instead been obsessed with a slightly different acronym and slightly different result.
We take a previously strong language model based only on boring LSTMs and get it to within a stone's throw of a stone's throw of state-of-the-art byte level language model results on \enwik.
This work has undergone no intensive hyperparameter optimization and lived entirely on a commodity desktop machine that made the author's small studio apartment far too warm in the midst of a San Franciscan summer\footnote{For those not in the know, San Francisco summers are generally not all that warm but I can't open my window as it overlooks an alley of questionable odors and unsavory noises.}.
The final results are achievable in plus or minus 24 hours on a single GPU as the author is impatient.
The attention mechanism is also readily extended to large contexts with minimal computation.
Take that Sesame Street.
\end{abstract}

\section{Introduction}

Language has been a thorn in humanity's side since we evolved a complex enough audio and graphics processing unit to grunt, let alone write cryptocurrency whitepapers or opinion columns.
Language has been found at the core of every human conflict in history, from World Wars (culinary and otherwise) to the Great Border Skirmish (2008) between you and your loud neighbour.
Many are of the opinion that language has redeeming features.
They claim (with scant evidence) that language could contain useful knowledge far beneath the surface wasteland of memes and colourful insults we usually see, just as life might theoretically be found deep under the ice of Jupiter's moon Europa.
Many fight against the homogenization of language by dividing and conquering as they did in the Tower of Babel era\footnote{Given the proliferation of non-human actors entering the linguistic ecosystem and the quality of their writing we may find natural language is already on this path - especially when it's hard to separate an inane machine's babble from an inane human's.} (see: Javascript frameworks).
Regardless of how you feel about language, a gambler would expect language to exist for at least a few more years and is thus a worthy field of study.

\subsection{Language is humanity's longest running program}

Perhaps the most surprising revelation of natural language processing and language modeling is just how richly embedded knowledge is in the standard text we write, the words we speak, and the ways in which we communicate.
History, wisdom, and computation are transcribed in the most boring of minutiae\footnote{The Vindolanda tablets show that parents can still embarass children by asking whether they have enough underwear and socks nearly two millenia later. Your social media posts will likely be the dataset used to produce billions of on-demand media experiences two millenia from now. You will get no royalties. Sorry.} whether we intend to capture it or not.

It's important to realize that language is far more than just human language as well.
Language spans every form of encoding: symbol, sight, or sound.
Many of the complexities and processes in the world can be rewritten into a language, whether the source is artificially produced (decoding the image of a CRT monitor via Van Eck phreaking) or naturally produced, as you might expect this text to be.

The concept of natural language extends far beyond the boundaries we might have put down for it in centuries past.
What we learn from trying to predict these patterns tends to be a rich source of knowledge, intended or otherwise.

\subsection{Direct motivation}

Language modeling is one of the foundational tasks of natural language processing.
The task involves predicting the $(n+1)^{th}$ token in a sequence given the $n$ preceding tokens.

In the natural setting it's that awkward date where they're trying desperately to finish your sentences to show you how smart they are - even if they're getting every guess wrong.
In the algorithmic setting it's the autosuggest that powers your phone, intelligently suggesting that your insult would be far more cutting if you mentioned ducking instead.


Why is language, and thus language models, such a rich source of knowledge?
Neural networks, plus or minus a few bajillion parameters, are theoretically capable of universal function approximation.
When you're asking a sufficiently complex neural network to approximate language you're asking it to approximate all the intricacies of the text, most of which you're likely not even consciously aware of.

Language models, at least as they stand, are not intelligent - but they do echo intelligence back at us.
The humans that created the vast datasets that define, describe, and expand our world are doing the intellectual work.
A language model, passing over enough text, is merely surfacing and connecting fragments of cached human computation.

Given the premise that language is humanity's longest running program, we now note that humanity have never made the best programmers.
Whilst spaghetti coding is the phrase used for unstructed and difficult to maintain code we also have spaghetti language.
Vast opportunities left to chance as we don't know how to organize ourselves\footnote{The early history of computing itself might have unfurled very differently if not for a chance meeting between von Neumann and Goldstine on a train platform in 1944, introducing von Neumann to ENIAC. Surely we can catalyze such chance occurrences better than stumbling across tweets on an adversarial non-chronological timeline? Why leave to dwindling chance in this information age?}.
Great works left undiscovered and uncontinued as the call stack return address was empty.
Knowledge hidden away, never to be referred to again until the garbage collector comes without mercy.
That's what I hope for language models to solve.

Here we'd traditionally discuss what exactly a token is.
A token is whatever you'd like it to be.
Language models have been shown to work on pixels, audio, characters, words, and so on.
Language modeling on text has traditionally focused on words, sub-words, and characters.

A point worth noting however is that by defining words or sub-words we're defining a major part of our language model's computation and structure.
An observation from the past few years of progress in machine learning: the more work humans put in to inflexibly defining a machine learning model the more likely we've accidentally handicapped it.


\section{Motivation}

\subsection{Alternate history and the genre of research fiction}

Imagine if we existed in a timeline where ``Attention Is All You Need'' \cite{vaswani2017attention} became the title of a hit pop song and the original auhors decided not to publish their work\footnote{Perhaps the reserchers give up on machine learning and instead decide to pursue a musical career themselves? Who am I to dictate what they do in this alternate timeline?
}.
What's next?
Do we expect research progress to stall until multi-head attention is eventually reconstructed in a secret underground bunker built to protect the last few surviving bright eyed researchers from the unwarranted attacks of Reviewer \#2?
This seems unlikely\footnote{Seriously - think of all the wonderful research domains that deep learning has now expanded into and then remember that for decades the phrase ``neural network'' was met with derision. Do you care that neural networks get stuck in local optima now?}.
Progress would likely be made in a different research area and we would expect to see the surrounding research areas flourish, akin to a bridge being built across a previously impassable river.

I played this scenario out in my head, fighting the varied cries of ``Give up on the LSTM \cite{Hochreiter1997LongSM}, it's dead technology, like the car phone or your Tamagotchi or your girlfriend's click wheel iPod or \ldots'' that I heard from many of my researcher friends, and the model described in this paper is the result.

To clarify, I'm entirely happy if this model fails, but why dismiss possibility out of hand?
Why crowd a single direction of progress like moths swarming a light bulb?

\subsection{Fighting the direction of research}

I have a belief, so far holding, that what may take a cluster to compute one year takes a consumer machine the next\footnote{\url{https://smerity.com/articles/2018/limited_compute.html}}.
This holds true not due to some mythical advancement of the technology in the preceeding year but simply as there are usually far more efficient ways to achieve something once we know it's possible.

Similar to alternative history and related to the long trend of Moore's law we can't be certain that efficiency will eventuate.
There is no guaranteed end state - much of this is instead an implicit goal set and targeted by the community.

If training and research aggregates around large compute models that are difficult to replicate then only large compute models that are difficult to replicate will stand to benefit from the continued research improvements of our field.
If minicomputers had faltered and mainframes had won in the late 20\textsuperscript{th} century we may well be talking about the inevitable dead end that was small form independent machines.

\subsection{A lone GPU}

Irrational as it seems I didn't want to use a cluster in the cloud somewhere, watching the dollars leave my bank account as I run various experiments.
All the work here was thus done on a single GPU.
Minor evaluation and back-of-the-envelope experiments for models were done using a secondary GPU whilst training continued on the first.

I'm not against using many GPUs, and feel I would use them efficiently, but have yet to find a cost effective way of doing so.
Friends primarily suggest using free credits on cloud services - but that doesn't seem sustainable.
As an independent researcher I fear losing access to compute, falling down the sheer cliff that I put myself on, having the magic carpet yanked from under my feet, having my carriage (re)turn into a pumpkin, and other related night terrors.
Perhaps I'm irrational and should grab the compute that I can get whilst I can get it - but I know that if I lose such compute it'll impact my morale and fighting spirit.
I am also not convinced yet that I can't do better with less which means invariably I'd be doing worse with more.

If I truly believe that "what may take a cluster to compute one year takes a consumer machine the next" then I should also be willing to live it - at least to some degree.
All my best work seems to come from being relatively low resourced and creative anyway.
That just also happens to be the right strategy to maintain a long runway and an open mind.
If I get a seed of possibility that works well and would flourish with more compute then the compute will be there waiting.

Sustainable compute is what I'm after.
Not just for myself but as I genuinely believe that machine learning should follow the history of mini-computers that got us each our own device in the first place.
Proper partnerships, not a smattering of free credits here and there, with entities that get value from the work performed.
Those parternships could be individuals or companies - it's more about the aligned values.
We need a Moore's Law for machine learning that encourages a minicomputer future, not a mainframe one.

\section{Model architecture}

\begin{figure}[t]
 \centering 
 \includegraphics[width=50mm]{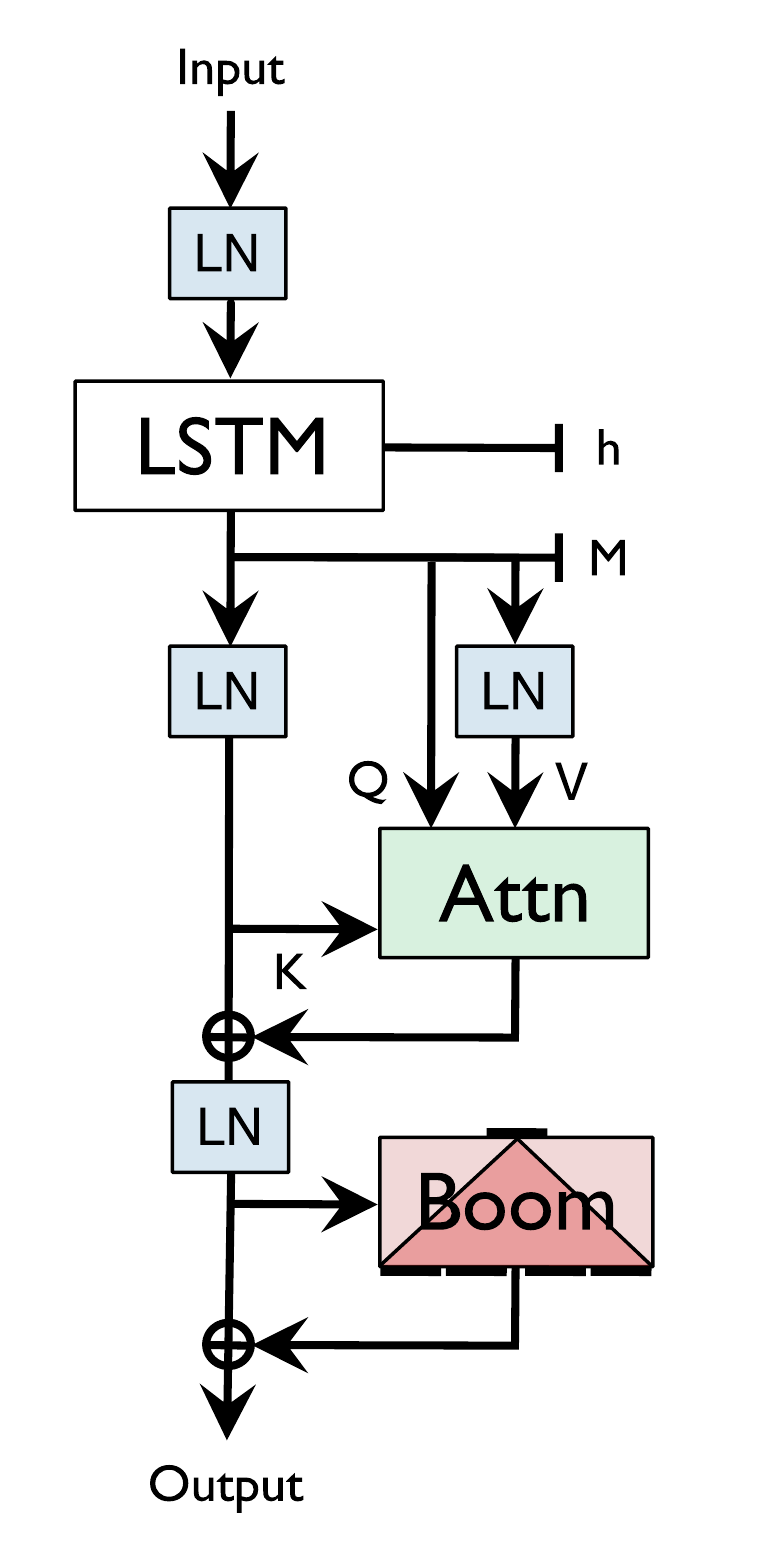}
\caption{The SHA-RNN is composed of an RNN, pointer based attention, and a ``Boom'' feed-forward with a sprinkling of layer normalization. The persistent state is the RNN's hidden state $h$ as well as the memory $M$ concatenated from previous memories.
Bake at 200$^{\circ}$F for 16 to 20 hours in a desktop sized oven.}
\label{fig:sharnn}
\end{figure}

The model architecture is an upgrade of the AWD-LSTM in \citet{merity2018regularizing}.
The code is a tad horrific.
I'll admit to that.
I know the codebase well enough that the cruft doesn't slow me down.
Like Han in the Millenium Falcon, it's a hunk of junk, but it's a hunk of junk that I know how to fly.

After this paper I plan on rebuilding the codebase from the ground up both as an educational tool for others and as a strong platform for future work in academia and industry.

The model consists of a trainable embedding layer, one or more layers of a stacked single head attention recurrent neural network (SHA-RNN), and a softmax classifier.
The embedding and softmax classifier utilize tied weights \cite{Inan2016,Press2016}.

The model uses a single head of attention, which is more along the lines of the Continuous Cache \citep{Grave2016} or Pointer Sentinel \citep{Merity2016}, and a modified feedforward layer similar to that in a Transformer, which I have referred to internally as a Boom layer.
I might as well suggest it be a Boom layer to you too given the amount of fun I've had saying it.
Why Boom?
We take a vector from small (1024) to big (4096) to small (1024).
It's really not that hard to visualize - use your hands if you need to whilst shouting "boooOOOOmmm".

\subsection{A simplified attention mechanism}

The attention mechanisms as used in many Transformer inspired architectures assume no sequentiality in construction and many complex attention heads - dozens per layer.
One might take issue with this as simple is better than complex.

\begin{figure}[t]
 \centering 
 \includegraphics[width=50mm]{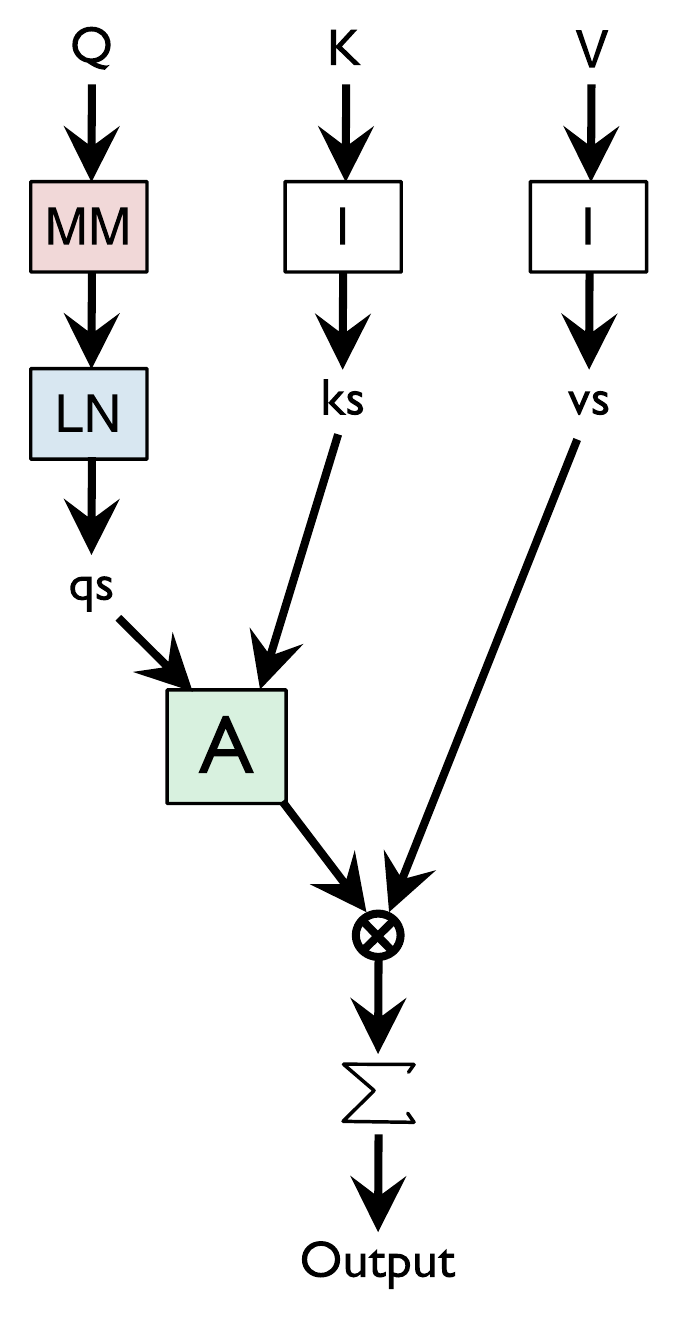}
\caption{
The attention mechanism within the SHA-RNN is highly computationally efficient.
The only matrix multiplication acts on the query.
The $A$ block represents scaled dot product attention, a vector-vector operation.
The operators $\{qs, ks, vs\}$ are vector-vector multiplications and thus have minimal overhead.
We use a sigmoid to produce $\{qs, ks\}$.
For $vs$ see Section \ref{sec:opslv}.
}
\label{fig:attn}
\end{figure}

The attention mechanism in this work has been simplified for two reasons.
First, does anyone have any freaking clue how many attention heads we need \cite{shazeer2019fast}?
Second, why are we putting so much work into the attention layer?
Are we sure about the benefit that this swarm of attention heads, each involving substantial compute, bring?

By having a single attention head I stopped running out of memory.
By having the memory written once and never modified (i.e. no big matrix multiplication each timestep), similar to that in the Continuous Cache and Pointer Sentinel model, we avoid a great deal of computation.
Life is simpler, roses smell better, wealth is better distributed across society, and you can extend your memory window indefinitely with minimal overhead.

\subsection{Boom layer}

The Boom layer is related strongly to the large feed forward layer found in Transformers and other architectures.
For whatever mad reason I decided to rearrange this a little.
The layer takes a vector of the form $v \in \mathbb{R}^H$ and uses a matrix multiplication with GeLU activation to produce a vector $u \in \mathbb{R}^{N \times H}$.
We then break $u$ into $N$ vectors and sum those together, producing $w \in \mathbb{R}^H$.
This minimizes computation and removes an entire matrix of parameters compared to traditional down-projection layers.


\section{Experiments}


Our work is primarily over the byte-level \enwik dataset though references the word-level used-to-be-large-scale WikiText-103 dataset for discussions on tokenization.

As the author of WikiText-103 it's also likely time to replace it with a new dataset.
This dataset will be based upon Wikipedia, once again, but will be intended as an Evergreen WikiText (EWT), both in the practical content it captures as well as the construction of the data itself.
If you'd like to help the Evergreen WikiText initiative at any level, be it as a human being with language skills, a researcher with hatred at my existing tokenization, a corporation with $+ve$ purpose, or a kind AGI, $\frac{d}{dx}$ Labs will accept your help.

\begin{table*}
\small
\begin{center}
 \begin{tabular}{@{} lrllr @{}}
 \toprule[1.5pt]
 Model & Heads & Valid & Test & Params \\
 \midrule
 Large RHN \cite{Zilly2016}& 0 & $-$ & 1.27 & 46M \\
 3 layer AWD-LSTM \cite{merity2018analysis} & 0 & $-$ & 1.232 & 47M \\
 T12 (12 layer) \cite{al2019character} & 24 & $-$ & 1.11 & 44M \\
 LSTM \cite{melis2019mogrifier} & 0 & 1.182 & 1.195 & 48M \\
 Mogrifier LSTM \cite{melis2019mogrifier} & 0 & 1.135 & 1.146 & 48M \\
 \midrule
 4 layer SHA-LSTM ($h=1024$, no attention head) & 0 & 1.312 & 1.330 & 51M \\
 4 layer SHA-LSTM ($h=1024$, single attention head) & 1 & 1.100 & 1.076 & 52M \\
 4 layer SHA-LSTM ($h=1024$, attention head per layer) & 4 & 1.096 & 1.068 & 54M \\
 \midrule
 T64 (64 layer) \cite{al2019character} & 128 & $-$ & 1.06 & 235M \\
 Transformer-XL (12 layer) \cite{dai2019transformer} & 160 & $-$ & 1.06 & 41M \\
 Transformer-XL (18 layer) \cite{dai2019transformer} & 160 & $-$ & 1.03 & 88M \\
 Adaptive Transformer (12 layer) \cite{sukhbaatar2019adaptive} & 96 & 1.04 & 1.02 & 39M \\
 Sparse Transformer (30 layer) \cite{child2019generating} & 240 & $-$ & 0.99 & 95M \\
 \bottomrule
\end{tabular}
\end{center}
\caption{
Bits Per Character (BPC) on \enwik.
The single attention SHA-LSTM has an attention head on the second last layer and had batch size 16 due to lower memory use.
Directly comparing the head count for LSTM models and Transformer models obviously doesn't make sense but neither does comparing zero-headed LSTMs against bajillion headed models and then declaring an entire species dead.
The hyper-parameters for the fully headed SHA-LSTM were used for the other SHA-LSTM experiments with zero tuning.
}
\label{table:enwik8}
\end{table*}

\subsection{Hutter Wikipedia Prize (\enwik)}

The Hutter Prize Wikipedia dataset \cite{hutter}, also known as \enwik, is a byte-level dataset consisting of the first 100 million bytes of a Wikipedia XML dump.
For our experiments, we follow a standard setup where the train, validation and test sets consist of the first 90M, 5M, and 5M characters, respectively.

Whilst I think this dataset needs an update given the data is from 2006 it has many positives.
First, the detailed history of traditional compression captured on a plaintext leaderboard is heart warming.
Second, the lack of preprocessing and thus the inclusion of Wikipedia markup has some interesting linguistic advantages.
Curious how a car name might be shortened?
Well, it mentions \textsc{the [[BMW M1|M1]] supercar}''.
What is Cobol? Well, \textsc{[[COBOL]] or [[APL programming language|APL]]}.
Ordo-vikki what? \textsc{[[Ordovician-Silurian extinction events|End Ordovician]]}.
How do you make a verb out of a chemical? \textsc{Ethylene can be [[chlorine|chlorinated]] to produce \ldots}.
The markup is rich with such examples.

\subsection{WikiText}
The WikiText-2 (WT2) and WikiText-103 (WT103) datasets introduced in \citet{Merity2016} contain lightly preprocessed Wikipedia articles with a closed vocabulary.
The vocabulary was constructed by splitting on spaces and punctuation  discarding all words with a count below 3.
The WT2 and WT103 datasets contain 2 million and 103 million words in the training set respectively.
Both have validation and test sets of 0.2 million composed of the same text.
As the Wikipedia articles are relatively long and are focused on a single topic, capturing and utilizing long term dependencies are key to models obtaining strong performance. 

\subsubsection{Wordpiece WikiText-103}

The WikiText dataset, as released by the foolish creator of that dataset, had a closed vocabulary.
That used to be a reasonable idea - only machine translation used wordpieces.

Wordpieces allow for a smaller model, better utilization of a model's parameters, and better compositionality within the language itself.
The utilization of a model's parameters are an important concept.
The larger a vocabulary the more difficult it is to select a given word (i.e. the issues addressed by approximate softmaxes and mixture of softmaxes) and the less likely it is for a given token's parameters to be accessed in a given context.

Due to the Zipfian nature of language distribution, most words won't occur in most contexts.
This means that, by extension, the parameters of rare words aren't of use to the model most of the time.
Such rare parameters also provide an easy opportunity for overfitting.
Thus not only do these tokens take up a lot of space but they're rarely used and when they are used can be a source of problems.


The WikiText-103 dataset was converted from the closed vocabulary of 267,735 tokens to a wordpiece vocabulary of 8000 using YouTokenToMe\footnote{\url{https://github.com/VKCOM/YouTokenToMe}}.

\section{Results and analysis}

\subsection{SHA-RNN on \enwik}

After hacking and slashing at the codebase to produce the SHA-RNN formulation above the results were quite pleasing.
Sadly I didn't hit state of the art results, at least as they stand today with minimal hyper-parameter exploration.
Yet these results would have been state of the art compared to many of the still evolving Transformer models of earlier papers.
More specifically I think they prove well that there should still exist competition and variety in the types of models that we put to our tasks - especially as language modeling now forms the basis of pretraining used by many NLP tasks across our field.

For the given parameter count the SHA-RNN model performs admirably.
LSTMs are forced to have their hidden state expressed primarily through recurrence, limiting their expressiveness.
The key, value, and positional information used by both the attention mechanism as well as later SHA-LSTM layers must also be encoded in the $h=1024$ dimensional vectors produced by the model at each layer.

Literally the night before submitting this paper I decided to run one more experiment - is it a single headed attention RNN with a single head or a single headed attention RNN with attention at each of the RNN layers?
Why not both?

Well, turns out the joke of ambiguity was on me.
A single head of attention gets almost all of the gains of the four layers of attention.
An additional advantage of the single headed SHA-LSTM is that each epoch took almost exactly $1800 \pm 1$ seconds (30 minutes) compared to the 4 headed SHA-LSTM which took 4020 seconds (67 minutes).
Due to lower memory usage the model was able to use a batch size of 16 rather than 8.
Training used 25 epochs with the last two at half learning rate.

That's what I mean about alternative research history.
What if we had this single single headed attention experiment two years earlier?
What if we as a field had spent our engineering efforts and experiment tokens on fast layer normalized block sparse LSTMs?
I'm not here to argue one way is right or wrong but simply that as a community we've gone a specific direction whether consciously intended or not.






\begin{figure}[t]
 \centering 
 \includegraphics[width=0.4\paperwidth]{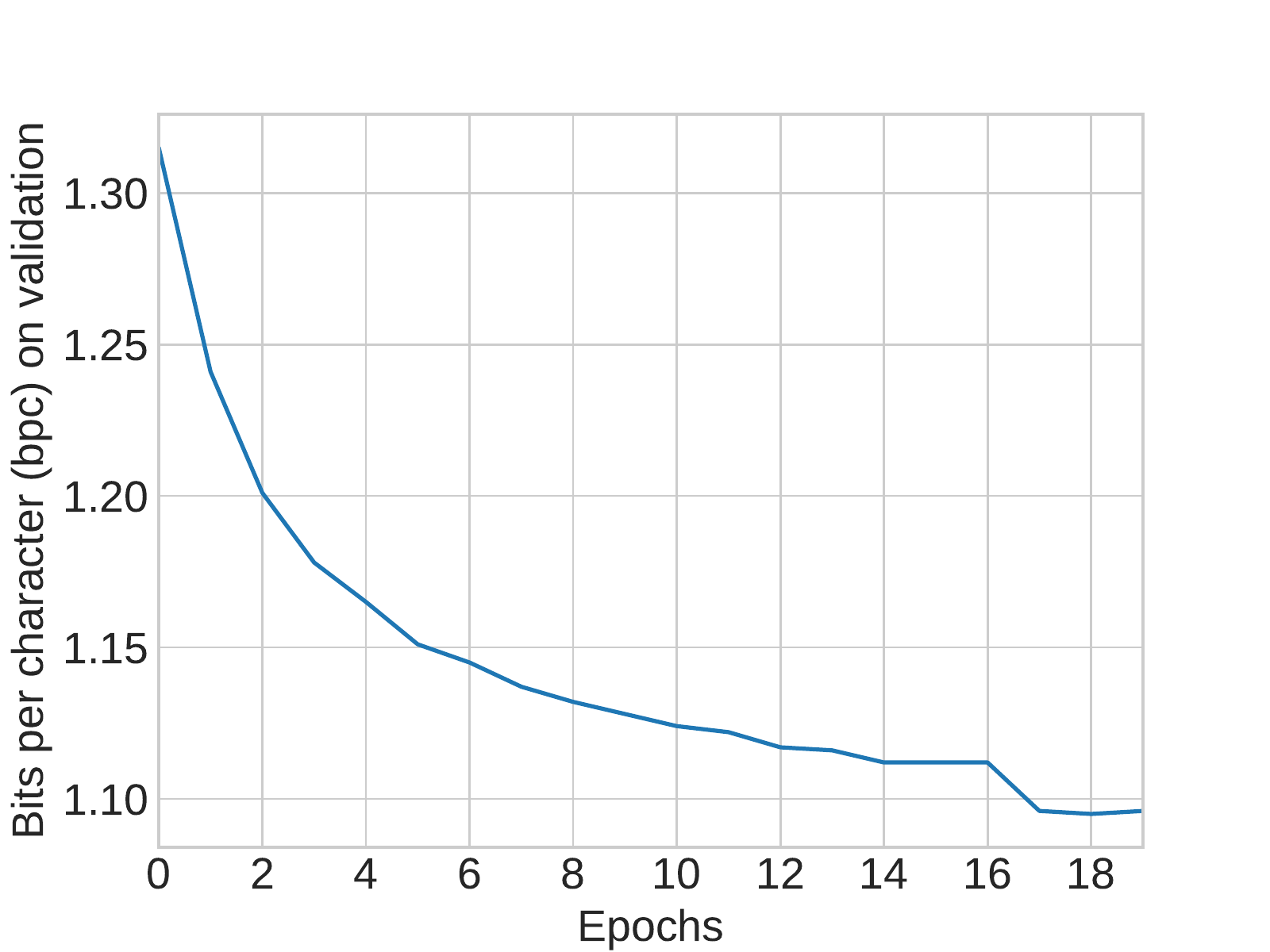}
\caption{Bits Per Character (BPC) reported on the \enwik validation set. Each epoch took approximately 4020 seconds (67 minutes). The learning rate was cut in half at epoch 16.}
\label{fig:bpcsharnn}
\end{figure}

\begin{table}
\renewcommand{\arraystretch}{1.1}
\begin{center}
 \begin{tabular}{@{} lrrrrr @{}} 
 \toprule[1.5pt]
 \midrule
 RNN Cell & \small{LSTM} \\ 
 Layers & 4 \\
 All hidden sizes & 1024 \\
 Input embedding size & 1024 \\
 Boom hidden size & 4096 \\
 Dropout (e/h/i/o) & 0/0.1/0.1/0.1 \\
 \midrule
 Optimizer & Min Trust LAMB \\
 Weight decay & $0$ \\
 BPTT length & 1024 \\
 Memory window size & 5000 \\
 Batch size & 8 \\
 Learning rate & 0.002 \\
 \midrule
 \bottomrule[1.5pt]
\end{tabular}
\caption{
Hyper-parameters for the language modeling experiment over the \enwik dataset.
Dropout refers to embedding, (RNN) hidden, input, and output.
Notice the lack of specifics in the hyper-parameters due to not using bonfires of compute.
\label{table:fullhypers}}
\end{center}
\end{table}

\section{Discussion}

\subsection{Tokenization attacks may break perplexity} 

\textit{Author note}: Initially this paper presented results on a wordpiece WikiText-103 with varying tokenizations.
The experimental formulation did not correctly factor in the normalization constant however and the wordpiece vocabularies were not constructed in a way that allows for proper comparison\footnote{Discussed in a tweet: \url{https://twitter.com/Smerity/status/1192252146126688256}}.
Future material will aim to provide a constrained experiment demonstrating the impact of varying tokenization.

When comparing language models trained on datasets with differing tokenizations a conversion process has traditionally been used.
Some papers have retokenized word level datasets, such as WikiText-103, using an invertible wordpiece tokenizer.
The resulting perplexities are then renormalized according to the number of tokens in each test dataset \cite{mikolov2012subword, hwang2017character}.

As an example, a formula exists to convert a bits per character (BPC) model to word level perplexity such that:
$$ppl = 2 ^{bpc \times \frac{N_c}{N_w}}$$
where $N_c$ is the number of tokens in the byte level dataset and $N_w$ is the number of tokens in the word level dataset.

Such a formula assumes equal entropy across the constituent tokens and does not factor in information introduced during the language modeling process (teacher forcing).

The distribution of entropy over characters is known to be non-uniform.
Experiments have traditionally shown entropy concentrated on the first few characters of traditional word segmentations in character level models \cite{elman1990finding}.

Invertible wordpiece tokenization aims to smoothly distribute entropy across tokens, further amplifying this difference in distributed entropy compared to word level models.



Almost any wordpiece tokenization will split prefixes and suffixes from various words for example and result in compositionally more balanced word fragments.
Mixed with teacher forcing, which is present at both training time and test time, this could have quite a profound impact.

This non-uniformity is exacerbated by modern components such as the attention mechanism which can effectively copy the occurrence of previously unknown sequences when correctly context switched \cite{al2019character}.
This rewards models which either allow for an early context switch to occur (i.e. the first character of a token) and which break high entropy tokens into a sequence of lower entropy tokens.

To take an extreme example, imagine you were trying to guess the password on my laptop (``lolcats'').
If you were trying to enter it on my laptop you'd only get feedback upon submitting the entire phrase, not on individual characters.

A timing attack in cryptography is a side channel attack where a timing signal (in our parlance ``tokenization'' of a temporal sequence of operations) leaks information.
As we're already using the SHA-RNN and there are crude comparisons to timing attacks in cryptography we will refer to this as a tokenization attack.
If we check our password a character at a time, stopping when a character doesn't match, we know how much of our candidate password is correct and can search through the remaining possibilities.

The situation in tokenization is worse as the teacher tells the attacker what the correct token actually was (``The password starts with an L, not a T'') and each token reveals far more information.
As noted earlier, natural language has an uneven distribution of entropy whilst a well crafted password should have near constant entropy throughout.

Important to note is that these tokenization attacks are a natural result of our language models accessing and relying on side-channel information that researchers may not be aware they're providing.
Deep learning is the ultimate spaghetti code\footnote{\url{https://twitter.com/Smerity/status/1174811071162376192}}.
Your model will hunt down the worst tricks, find and exploit every edge case, proceed to make a mess of it, and then trick you into thinking it's working.
Tokenization attacks are but one example.

\subsubsection{Examples of tokenization attacks}

Any time you need to guess between ``specialized'' and ``specialised'', occurring 1091 times and 430 times in the training data respectively, you're going to get a far larger perplexity for a word level model than if your wordpiece model breaks them down into 
\textbf{\textcolor{red}{special}}ize\textbf{\textcolor{red}{d}} or similar.

The same is true when copying words and named entities, strongly assisted by attention mechanisms.
If we're answering ``Who was killed? Dumbledore or Gandalf?'' with a word-level model then we have only one guess.
A wordpiece model has a massive advantage however as entropy may be loaded mostly on the first token (i.e.
\textbf{\textcolor{red}{D}}umble\textbf{\textcolor{red}{dore}}).

\subsection{Improvements: easy in theory, hard in practice}

Whilst we are in a relative golden era for deep learning frameworks we still have some fundamental limits to novel endeavour to consider.

Custom high performance components are still hard to write.
Without fear of slicing an artery open on bleeding edge features it's still difficult to write a fast LSTM kernel for example.
There exist no tuned and widely supported implementations with layer normalization, block sparse weights, and so on.
NVIDIA have not released a new variant of the LSTM in their cuDNN package for many a SotA.

Mixed precision (i.e. fp16) is still a nightmare even with a relatively lovely library or framework to deal with many of the issues.
The NVIDIA Apex library deserves a hand-written thank you note from me but it is still a thin veneer of decoration on top of a bubbling vat of acid.

For these reasons, as well as the primary author being both lazy and ineffective, the implementation of SHA-RNN could be far more efficient, both at a low level and high level.

Whilst that's a minor tragedy in this paper, the broader tragedy is related to the alternate history discussion from earlier.
The successful methods get prioritized in both academic improvements as well as practical.
How much does that dictate the progress in our field?
How can we make it easier for wild experiments to still be well supported by the intellectual and practical tooling of our field?

If I were using a layer normalized LSTM\footnote{I have written custom RNNs before but they're usually a paper worth of effort themselves. It's far easier to rely on existing black box implementations, such as NVIDIA's, which have been optimized than roll your own. Rolling your own usually results in either a speed drop or nights of muttering to the $\frac{d}{dx}$ gods.}, which has been shown to work well at the byte level compared to Transformers in the compression setting \cite{bellard2019lossless}, would we have seen a larger jump?
This is an important and even question but in daring to ask it you're not just fighting the trend of research but also the trend of engineering.
It's exhausting enough fighting one direction let alone both.

\subsection{Minimum Trust LAMB}

To improve convergence we use the LAMB optimizer \citep{you2019reducing}.
The LAMB optimizer, having been used primarily on residual layer normalized networks such as the Transformer architecture, appear to fail to converge for models such as the SHA-RNN when layers are not residual.
To handle this we introduce a minimum trust to the LAMB optimizer to ensure a minimum amount of progress is made on each pass\footnote{\url{https://github.com/Smerity/pytorch-lamb/commit/704f733c83c18fc5f3c01f085b5beb38043b38af}}.
For all experiments we set the minimum trust ratio to $0.25$.

\subsection{Over-paramaterizing static learned vectors}
\label{sec:opslv}

\subsubsection{The truth}

During development I made mistakes.
Many mistakes.
So very many mistakes.
What I find far more concerning is that in my history as a professional neural network nudger I've made numerous mistakes that have ended up helping me in the end.
In this section I shall first describe the mistake and then write it as if I had meant to do it the whole time with my intuitions as to why it helps.
The last part is especially important if you're planning to submit your academic paper to conferences\footnote{This suggestion is mostly a joke but the fact that this conduct is essentially expected in our academic field is also a joke. Your reviewers would be happier if you waved your hands around with equations pretending you knew what was happening rather than admitting you don't know and positing an educated guess.}.

For the attention pass I wanted to give the model the opportunity to disregard certain parts of the vector space.
This would be important as parts of the vector will likely end up reserved for the LSTM (local information) whilst other parts of the vector will likely end up being reserved for long range attention (global information).
The easiest thing to do is to multiply the given vector $o$ with a sigmoid (i.e. ranging from 0 to 1) mask $m$.
Awesome.
$o \cdot m$.
Seems easy.

Well, I blundered and accidentally passed $m$ through a QRNN - basically a fast version of the LSTM if you've not been keeping up with my Salesforce era research.
This happened for quite some time.
Big oops.
The good news is that it should essentially be a null op!
A crazy big matrix multiplication of $m$ followed by a tanh activation where only parts of the result are exposed thanks to another big matrix multiplication followed by a sigmoid!
In the end that's just a boring vector as the output - so if you simply set $m = \textit{QRNN}(m)$ then you have the exact same result.

Well, almost, as this holds true for static models, but if I tried training it without this weird QRNN initialization the results were worse.
Wut?
Wut indeed my astute reader.
I have theories but I never really quantified all of them and only have guesses.
In grand academic tradition however let's pretend I meant to do it the entire time.

\subsubsection{The polish}

In many complex deep learning systems there exist static learned vectors of the form $v \in \mathbb{R}^H$ where $H$ is the model's dimensionality.
These static learned vectors shift during training but are fixed in production.
They interact with millions or billions of dynamically produced vectors by their use in initial hidden states, gating mechanisms, and so forth, aiming to optimize a particular objective.
This results in a complex optimization landscape that low dimensional and historyless\footnote{Historyless meaning that this vector can't keep track of which past vector values were useful and which might have been problematic. All the history is removed immediately when the values are changed.} vectors are ill equipped to handle.

If we had fixed inputs (i.e. we could fix the dynamically produced vectors that these vectors interact with) then learning the optimal static vector could be achieved by optimizing over the entire dataset.
With a small batch size we would expect the value of our static learned vectors to potentially deviate wildly during training, requiring many epochs and a decaying learning rate before converging on an optimal stable value across the entire dataset.

In complex deep learning systems however we do not have fixed inputs.
During training our learned vectors interact with the dynamically produced vectors changing the trajectory of optimization for the entire system.
As the static learned vector has no aspect of history any previous optimizations are immediately lost, amplifying the potentially wild deviations seen during training and preventing stable training.

By over-parameterizing the static learned vectors we can provide an aspect of history to these low dimensional entities.
Rather than modifying the values destructively the model now has the option to change the eventual value by shifting the weights within the model, the initial vector, or the sigmoid gates.

For a simple example imagine there were two optimal values during optimal training - the zero vector $z$ and the mask vector $m$.
If we simply learned the correct sigmoid gates $g$ such that the output is $m \cdot g$ then the model can jump (on each model weight update) between outputting a zero vector and outputting the mask vector $m$ without having to relearn the value from scratch.
The history is stored in $m$ and protected by the model only changing $g$.
This is a vast oversimplification for complex models but the intuition should follow.

For the value of $vs$ in Figure \ref{fig:attn} an over-parameterized component was used.
This component took as input a single vector $v \in \mathbb{R}^H$ and produced a forget gate $f \in \mathbb{R}^H$ and candidate $c \in \mathbb{R}^H$ through a matrix multiplication and a sigmoid and tanh activation respectively:
$$vs = \sigma(W^f v) \cdot tanh(W^c v).$$
After training we can remove all of these parameters by replacing $vs$ with the now static output of the above equation.
Thus these parameters only exist during training and do not count toward the final model's parameter count.




\subsection{So you're saying Transformers are useless?}

Wait, really?
I didn't say anything close to that!
There are many situations where having eighteen bajillion attention heads is likely the best choice.
Any situation that involves minimal or uncertain sequentiality suddenly leans heavily to a multi-headed attention approach.
Yet are we certain we want to go all in on that direction?
Is it possible we began paying a complexity tax that was never immediately clear?

If anything my approach simply suggests there are many different paths, all viable, to solving the tasks we face.
No single architecture is likely going to take over the world indefinitely or be the optimal solution to all our woes.

For the task of language modeling I submit the SHA-RNN as an antidote to repeated and excessive Transformer exposure.

Perhaps we were too quick to throw away the past era of models simply due to a new flurry of progress.
Perhaps we're too commited to our existing stepping stones to backtrack and instead find ourselves locked to a given path.

Whether or not we're asking these questions consciously as a community we're still actively making these decisions.
Each decision we make, and each simpler solution we discard and build prejudice against, is likely to determine our path.

\section{Conclusion}

The Single Headed Attention RNN (SHA-RNN)
achieves strong results with next to no hyper-parameter tuning.
Even if this architecture doesn't catch on it still serves to show that the interaction between model performance and attention heads are not as clear as we might have guessed.
We have hope that LSTMs are not yet dead.
Whether that remains true is as much an engineering question as a research one.

If the SHA-RNN does catch on it could become a basis for model distillation techniques and/or be trained to take on a frightening battlefield of deranged muppets and transformable cars in the near future.

We also introduce the concept of a tokenization attack and note why varying tokenization schemes may prevent direct comparisons between models when factoring in teacher forcing.


\section{Acknowledgement}

This work was almost entirely run on a lone Titan V GPU donated by NVIDIA to me some time ago.
At a glance the Titan V appears to be composed of more gold than the average randomly generated world in Minecraft contains.

I would also like to commend the Adaptive Transformer \cite{sukhbaatar2019adaptive} as it's one of the few Transformer based architectures that is trainable to strong results on a single GPU in a matter of hours.

Thanks to Bryan McCann, Yaroslav Bulatov, Otavio Good, and others in long form discussions regarding renormalization of perplexity scores when varying tokenization.

The author has also moved to a one bedroom apartment in San Francisco, removing themselves from proximity to the alley of questionable odors and unsavory noises.

\bibliography{example_paper}
\bibliographystyle{icml2018}

\end{document}